\documentclass[num-refs]{wiley-article}
\usepackage{anyfontsize}  
\usepackage{svg}

\usepackage{hyperref}
\usepackage{siunitx}
\usepackage[justification=centering]{subfig}

\usepackage{color}
\usepackage{ifthen}

\providecommand\todonotesactive{false} 

\ifthenelse{ \equal{\todonotesactive}{true} }{
  \newcommand\TODO[1]{\if\relax\detokenize{#1}\textbf{\textcolor{red}{?}}\else\textbf{\textcolor{red}{#1}}\fi} 
  \setlength{\marginparwidth}{1.2cm} 
  \usepackage[colorinlistoftodos, textsize = small]{todonotes}
}{
  \newcommand\TODO[1]{} 
  \usepackage[disable]{todonotes} 
}

\newcommand\CHANGED[1]{\textcolor{blue}{#1}}

\papertype{Original Article}
\paperfield{Journal Section}

\title{Addressing the Challenges of Loop Detection in Agricultural Environments}

\author[1]{Soncini, Nicolas}
\author[2]{Civera, Javier}
\author[1]{Pire, Taihú}

\affil[1]{CIFASIS (CONICET-UNR), Rosario, Santa Fe, Argentina}
\affil[2]{I3A, Universidad de Zaragoza, Zaragoza, España}

\corraddress{Nicolas Soncini, Universidad Nacional de Rosario, Rosario, Santa Fe, 2000, Argentina}
\corremail{soncini@cifasis-conicet.gov.ar, jcivera@unizar.es, pire@cifasis-conicet.gov.ar}

\fundinginfo{
Consejo Nacional de Investigaciones Científicas y Técnicas, Grant/Award Number: PIBAA No. 0042;
Universidad Nacional de Rosario, Grant/Award Number: PCCT‐UNR80020220600072UR;
AGENCIA I+D+i, Grant/Award Number: PICT 2021‐570;
Spanish Government, Grant/Award Numbers: PID2021127685NB‐I00, TED2021‐131150B‐I00;
Aragón Government, Grant/Award Number: DGAT45\_23R
}

\runningauthor{Soncini, Nicolas et al.}

\begin{document}

\thispagestyle{empty}
\addtocounter{page}{-1}
\setlength{\parskip}{\baselineskip}
\setlength{\parindent}{0pt}

{
\centering
This paper has been accepted for publication in \textit{Journal of Field Robotics.}\\
This is the pre-peer reviewed version of the following article:

 Soncini, N., Civera, J. and Pire, T. (2024) Addressing the challenges of loop detection in agricultural environments. Journal of Field Robotics, 1–10. \url{https://doi.org/10.1002/rob.22414}.

which has been published in final form at \url{https://doi.org/10.1002/rob.22414}. This article may be used for non-commercial purposes in accordance with Wiley Terms and Conditions for Use of Self-Archived Versions.

}

\begin{frontmatter}
\maketitle

\begin{abstract}
While visual SLAM systems are well studied and achieve impressive results in indoor and urban settings, natural, outdoor and open-field environments are much less explored and still present relevant research challenges.

Visual navigation and local mapping have shown a relatively good performance in open-field environments. However, globally consistent mapping and long-term localization still depend on the robustness of loop detection and closure, for which the literature is scarce.

In this work we propose a novel method to pave the way towards robust loop detection in open fields, particularly in agricultural settings, based on local feature search and stereo geometric refinement, with a final stage of relative pose estimation.
Our method consistently achieves good loop detections, with a median error of $\SI{15}{\cm}$.
We aim to characterize open fields as a novel environment for loop detection, understanding the limitations and problems that arise when dealing with them.
Code is available at: \url{https://github.com/CIFASIS/StereoLoopDetector}

\end{abstract}
\end{frontmatter}

\ifthenelse{ \equal{\todonotesactive}{true} }{
  \listoftodos
}

\section{Introduction} \label{sec:introduction}
In the context of mobile robotics, loop detection refers to the automatic recognition of a re-visit to a previously traversed or known position~\cite{tsintotas2022revisiting}. This task is a crucial piece in the problem of Simultaneous Localization and Mapping (SLAM), that aims to map an unknown environment and localize the robot on it at the same time~\cite{cadena2016past}. Loop detection allows a SLAM pipeline to correct the drift caused by exploratory trajectories, thus improving the estimated robot trajectory and map.

Visual sensors are commonly part of the suite of sensors on a mobile autonomous robot. In outdoor robotics applications, in particular, there has been significant research on developing visual representations that are robust to changes in viewpoint, climate conditions, lighting conditions and even structural changes \cite{pire2023experimental}. More recently, such research has focused on learning such visual representations from data. Visual Place Recognition (VPR)~\cite{schubert2023visual} learns such representations by a retrieval formulation of the problem, in which given a query image the goal is to find the $k$ most similar images in a dataset of geolocated places. This is typically done by extracting the $k$-nearest neighbours in the representation space. Recent VPR works include ~\cite{chen2017hybridnetvpr,zaffar2020cohog, hausler2021patchnetvlad,berton2022rethinking,ali2023mixvpr,berton2023eigenplaces,zhu2023r2former,izquierdo2024optimal}. It should be noted, however, that loop closure typically includes the estimation of the relative transformation between the query and the retrieved images, and in this sense it is different from VPR.

The literature on the specific topic of loop detection is scarce, as it is usually part of, or associated to, a full SLAM pipeline. One can find methods that use handcrafted image features aggregated into a Bag-of-Visual-Words (BoW) with slight modifications, such as \cite{pire2017sptam,campos2021orbslam3}.
Loop closure has also been addressed for other sensor modalities, for example LiDAR~\cite{shan2020liosam,li2023semantic} and radar~\cite{hong2020radarslam}. However, typical benchmarks for SLAM and place recognition are limited to indoor~\cite{burri2016euroc} or autonomous driving~\cite{geiger2013vision} scenarios and ignore many other application domains.

Agricultural scenes, which are the object of this paper, present unique challenges that do not appear in urban or semi-urban environments, the most critical one being that visual features are scarce and repetitive \cite{cremona2022evaluation}. 
In a monoculture field there are vast aliased areas, covered by the same crop and at the same growth stage, with few or none locally salient elements and a high degree of inter-frame appearance changes due to illumination and weather \cite{pire2019rosario}. 
The most discriminative features tend to be distant structures near the horizon, with small image sizes and little geometric information. The rigidity assumption, typical in other domains, does not hold due to phenomena such as wind or rain, and the crop growth. Loop closure in agricultural domains is indeed under-addressed due to such challenges, together with the small number of of agricultural and open field datasets with accurate ground-truth annotations.

In this paper we present the first loop detection method which is specifically tailored for agricultural environments. The proposed method begins by a BoW search~\cite{dbow2012galvezlopez} to obtain rough loop candidates, which are then expanded with temporally close frames to select a best matching candidate, and finally verified by a geometry consistency check estimating the relative transformation between the stereo images. 
Our experiments quantify the performance that can be achieved by our method. After that, we conclude by characterizing agricultural environments in an attempt to continue improving visual localization for these open field environments.

The sections are organized as follows.
Section~\ref{sec:related} presents related work on Loop Detection and localization in agricultural environments. 
Section~\ref{sec:method} describes how our proposed method is designed.
Section~\ref{sec:results} shows results of the method tested on two agricultural datasets and discusses the results and the problem of loop detection on these environments.
Finally in Section~\ref{sec:conclusions} we present our conclusions and what we need to continue working on to achieve robust localization in unstructured environments such as these.

\section{Related Work} \label{sec:related}
We detail in this section methods that cover work related to loop detection, be it directly mentioned as such or not.

Within the SLAM literature, the majority of the best performing pipelines~\cite{pire2017sptam,qin2019vinsfusion,campos2021orbslam3,leutenegger2022okvis2} use variants of BoW~\cite{dbow2012galvezlopez} for loop closure. 
Several other loop closure methods can be found in the literature, such as FAB-MAP~\cite{cummins2008fabmap} with solid probabilistic foundations, or methods based on global image descriptors such as GIST~\cite{murillo2012localization} and binary versions of it~\cite{sunderhauf2011brief}, or Randomized Ferns~\cite{glocker2014real}, this last one used in ElasticFusion~\cite{whelan2016elasticfusion}. However, the practical advantages of a BoW representation and the accuracy of the relative motion from local features have made this approach the predominant one.

The loop detection problem based solely on visual information is a particular case of the area studied by Visual Place Recognition (VPR), where images are not required to be taken sequentially.
In \cite{garg2021whereisyourplacevpr} VPR is defined directly as "[...] a comparison of visual data, observed from same or different physical locations with same or different viewpoints." in which they claim it has to posses "[...] the ability to recognize one’s location based on reference and query observations perceived from overlapping field-of-views.", which will be key to discuss afterwards its shortcomings on detecting places in agricultural settings.
Among the methods that address the VPR problem, those based on deep networks stand out, such as NetVLAD \cite{arandjelovic2015netvlad}, Chen et al. \cite{chen2017hybridnetvpr} and SuperPoint \cite{detone2018superpoint}, which make use of convolutional networks, and SMART \cite{pepperell2014smart}, Chen et al. \cite{chen2014convnetvpr} and SeqNet \cite{garg2021seqnet} which make use of sequence information. 
In \cite{masonecaputo2021deepvprsurvey} a review of recent literature on VPR with deep networks is presented, however, works such as \cite{sunderhauf2015convnetperformance, zeng2018AS} show that variations in the point of view or changes in the lighting and structure of the environment continue to present problems even for these systems, which is why work is still required in this field.

Among the work on loop detection and closing are \cite{mcmanus2014shadydealings} which seeks to be resilient to lighting changes by using lighting-invariant images, \cite{naseer2014localizationacrossseasons} which builds a temporal association graph and generates multiple hypotheses to close a cycle even when faced with seasonal changes, and \cite{sattler2018benchmarking6dof} which performs benchmark work evaluating various VPR algorithms against lighting changes (day/night), changes in capture conditions (overexposure, blur), and changes in geometry (vegetation).
Our work is based on \cite{dbow2012galvezlopez} which uses a pipeline of local keypoint extraction, a hierarchical BoW search, and temporal and geometrical checks to perform fast loop detection on monocular images.

Recent literature has focused on distinctive environments in cities~\cite{warburg2020mapillary} or from a train~\cite{olid2018singlevpr} and primarily on the issues of lighting and appearance changes~\cite{torii201524} and only recently on view changes~\cite{vallone2022danish}.
We have found no work that covers loop detection on agricultural environments and believe that we're the first ones to take on this problem on an environment severely affected by perceptual aliasing, far away horizon elements, severe climate and illumination changes and a need for precise localization for autonomous robotics.

\section{Proposed Method} \label{sec:method}
We propose a method based on stereo visual information comprised of an initial similarity search and further temporal and geometric refinement, inspired by the method proposed in \cite{dbow2012galvezlopez}.
\CHANGED{We show our process using similar notation for comparison.}

\begin{figure}[!hbt]
    \centering
    \includegraphics[width=\textwidth]{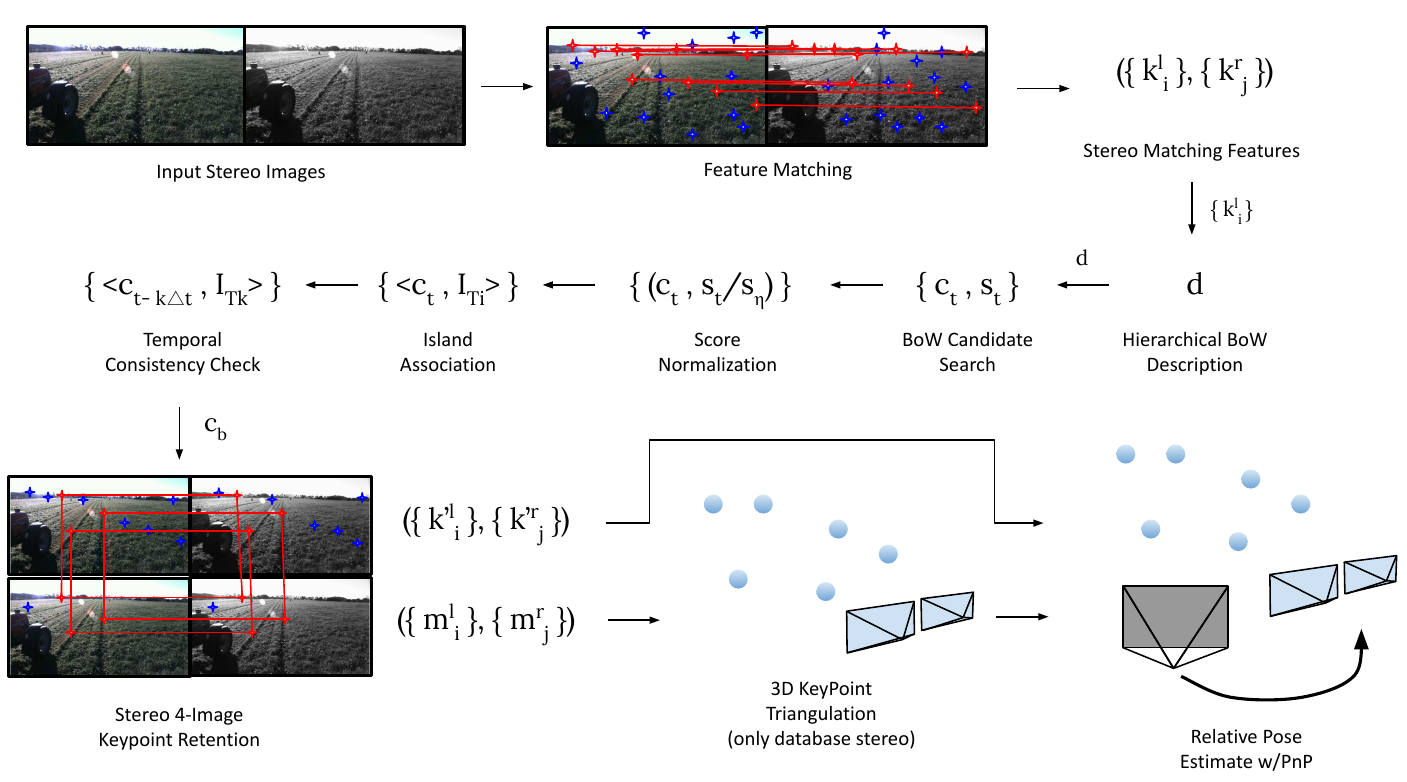}
    \caption{Overview of our loop detection pipeline, where $k$ are keypoints for the query image, $d$ is a descriptor of the left query image, $s$ are the similarity scores of the bow search, $I$ are the island sets, and $m$ are the keypoints for the best matching image.}
    \label{fig:sloopdetector_flowchart}
\end{figure}

As shown in Fig.~\ref{fig:sloopdetector_flowchart}, our method starts with an input of two grayscale images forming a calibrated stereo pair taken at time $t$.
Here we assume that the images are undistorted and rectified, we indicate information related to the left and right images with the letters $l$ and $r$ respectively.
Each stereo frame is processed individually to extract a set of salient visual features, and only the features that can be triangulated in the stereo pair are kept, which we symbolize with the sets $K^l$ and $K^r$ respectively, where we assume that the keypoints $k^{l}_i \in K^l$ and $k^{r}_i \in K^r$ are stereo matching keypoints for all $i = 1...n$, where $n$ is the number of matched keypoints.
We then generate a global image descriptor for the left stereo image $d_t$ using a hierarchical BoW~\cite{sivic2003videogoogle, nister2006scalablevoctree}  from the previously stereo-matched visual features.
This allows us to search very efficiently through a database of images of previous places and extract a set of potential loop closure candidates ${(c_t, c_{t_1}), (c_t, c_{t_2}),...}$, where $t_k$ represents an image taken at time $k$.

As originally proposed in \cite{dbow2012galvezlopez}, we apply several consistency tests to our set of potential loops.
The first one consists on a normalization of the similarity score based on the image similarity with the previous image on the sequence, that is, given a similarity score $s(c_t, c_{t_k})$ given by the BoW search, we normalize the similarity value $s(c_t, c_{t - \Delta t})$ giving us a new similarity score function 
\[
    \eta(c_t, c_{t_k}) = \frac{s(c_t, c_{t_k})}{s(c_t, c_{t - \delta t})}
\]
To prevent competition between temporally close images, the candidates are grouped on ``islands'' $I_{Ti}$ with images that are temporally close, that is, whose timestamps are close to each other by a preset maximum difference, forming the sets of timestamps $Ti = {t_{n_i}, t_{m_i}}$.
The islands are now re-ranked by a score $H$:
\[
    H(c_t, I_{Ti}) = \sum_{j=n_i}^{m_i} \eta(c_t, c_{t_j})
\]
and the island with the highest score continues to a temporal consistency check where the match $(c_t, I_{Ti})$ has several of it's previous images $c_{t - k\Delta t}$, for a predefined $k$, checked so that the islands $I_{Tk}$ they create have times close to overlap with times of the island $I_{Ti}$, as used in \cite{dbow2012galvezlopez}.
The final candidate $c_b$ is one belonging to the island $I_{Ti}$ that maximizes the scoring function $\eta$.

We then grab the candidate stereo image and query stereo image keypoints and filter those that match for all four images with an exhaustive matching algorithm, this leaves us with a pair of sets of filtered keypoints for the query stereo image $(K'^l, K'^r)$ and another for the matching stereo image $(M^l, M^r)$.
The filtered keypoints from the candidate image are triangulated to obtain 3d positions for all of them, and we filter them with maximum and minimum distance thresholds, making sure that we also filter the corresponding matching keypoints from the query image.
From the filtered candidate's 3d points and the filtered query left image keypoints we perform a Perspective-n-Point to obtain the relative position of the query image with respect to the candidate pose $P_{c_t, c_b}$, being this the output of our pipeline.

Our pipeline can be configured with several parameters related to the image processing (feature extractor, descriptor and number of features retained), the dataset (minimum temporal distance, amount of temporally consistent matches), and the geometrical check (minimum and maximum triangulated points distance, minimum numbers of features to use for PnP, matching function to perform), for some of which we perform ablations in Section~\ref{sec:ablations}.

\section{Experimental Results} \label{sec:results}
\subsection{Datasets}
Before detailing the datasets we use for the evaluation, it is worth pointing out the scarcity of open-field and agricultural datasets with positional ground-truth and effective loops.
We can observe a huge progress on indoors and urban dataset collection, fueled by their use in mixed reality and autonomous driving applications respectively. This allows different pipelines to be benchmarked in different conditions within such domains. However, the same cannot be said for agricultural settings.
We therefore make intensive use of the existing datasets that comply with our requirements (video recordings, accurate ground-truth position, agricultural setting).
This is a call to the scientific community working on these types of problems to expand the existing data to this ubiquitous task that is agriculture, which is as essential, if not even more, than any other.

\begin{figure}[!hbt]
    \centering
    \subfloat[]{
        \includegraphics[width=.32\textwidth]{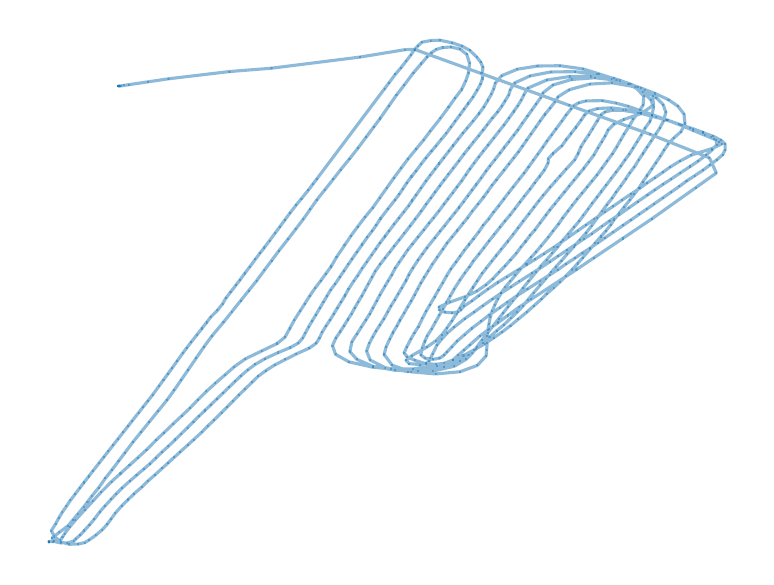}
    }
    \hfill
    \subfloat[]{
        \includegraphics[width=.32\textwidth]{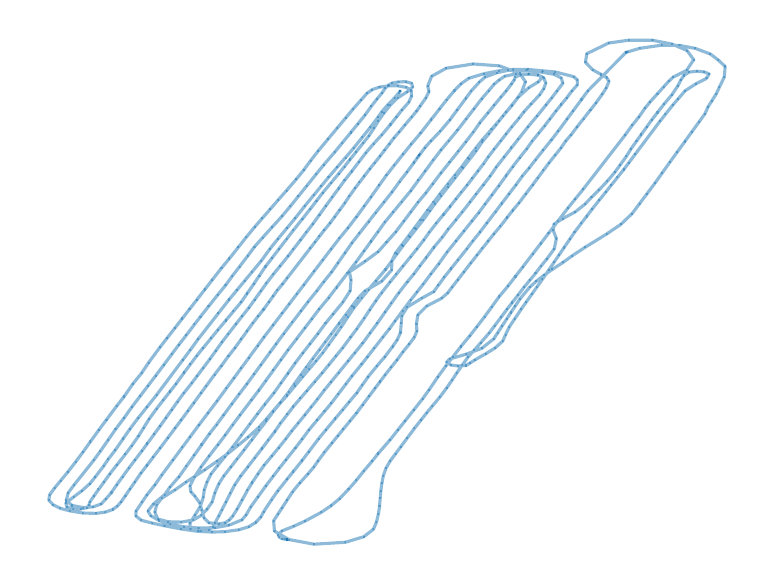}
    }
    \hfill
    \subfloat[]{
        \includegraphics[width=.32\textwidth]{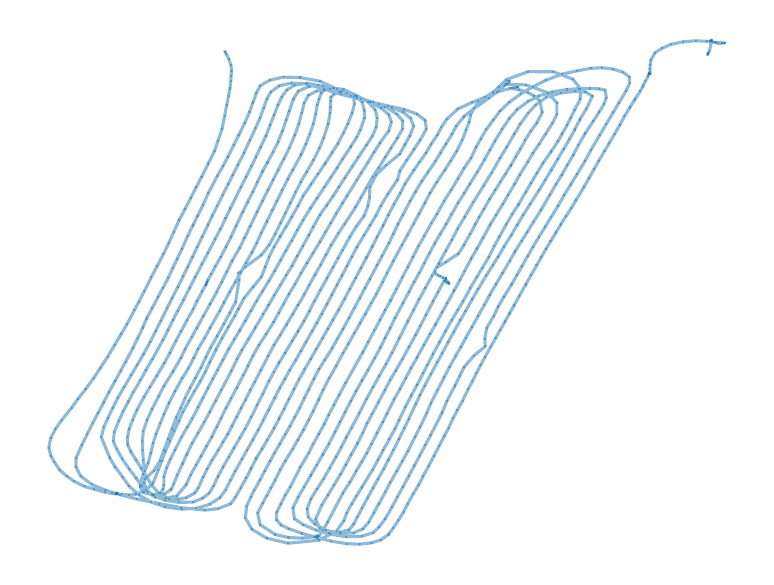}
    }
    \caption{Planar projection of the GPS-RTK trajectories for the three FieldSAFE sessions.}
    \label{fig:fieldsafe_trajectories}
\end{figure}

Specifically, we perform our tests on the ``FieldSAFE''~\cite{kragh2017fieldsafe} dataset, which was collected by a set of sensors mounted on a tractor running in an agricultural setting in Denmark in the year 2016.
It was conceived to capture data for object detection on agricultural settings, but it contains all the sensor data necessary to test a place recognition system, as the tractor system is equipped with the following sensors:
\begin{itemize}
    \item GNSS: it utilizes a \textit{Trimble BD982 GNSS}\footnote{\url{https://oemgnss.trimble.com/product/trimble-bd982/}} which provides us with a very accurate position estimate of the vehicle which we can use to derive ground truth place recognition.
    \item Web Camera: it's mounted with a \textit{Logitech HD Pro C920}\footnote{\url{https://www.logitech.com/en-eu/products/webcams/c920-pro-hd-webcam.960-001055.html}} from which we can test our methods for commercially available cameras.
    \item 3D Stereo Camera: it also carries a \textit{Multisense S21
    CMV2000}\footnote{\url{https://docs.carnegierobotics.com/S21/index.html}} which provides a stereo pair and depth information which we can use to test more advanced methods.
\end{itemize}
The FieldSAFE dataset comprises 5 different sessions, for which the robot performs a series of trajectories inside an agricultural field. We make use of 3 of the 5 sessions that showcase the most run-time and overlapping trajectories: ``Dynamic obstacle session \#1'', ``Dynamic obstacle session \#2'' and ``Static obstacle session \#2'', whose trajectories are shown in Fig.~\ref{fig:fieldsafe_trajectories}.
These selected trajectories have a duration of over $\SI{21}{\minute}$, $\SI{22}{\minute}$ and $\SI{26}{\minute}$, and a traveled distance of over $\SI{3}{\km}$, $\SI{4}{\km}$ and $\SI{4.5}{\km}$ respectively.

\subsection{Results}
Our implementation extracts $2000$ ORB keypoints for each image of the stereo pair, retains only the ones that match between the stereo images (via brute force matching) and adds them to the hierarchical BoW via a predefined vocabulary trained in the \emph{Bovisa 2008-09-01}~\cite{bonariri2006rawseeds,ceriani2009RAWSEEDS} dataset and available along the ORB-SLAM3 implementation~\cite{campos2021orbslam3}.
We set the normalized scoring threshold to $0.3$, and the temporal consistency to agree with $5$ previous matches.
A loop candidate is dropped if at any stage (stereo matching, stereo pair matching or PnP) we have less than $20$ image features.

The following parameters are dataset dependent, these are the ones we empirically found work best for FieldSAFE. We do not consider images as candidates for loop detection unless $\SI{20}{\second}$ passed between them, and we set the minimum and maximum distances to consider stereo triangulated points to $\SI{0.4}{\meter}$ and $\SI{50}{\meter}$ respectively due to the stereo camera baseline.

\begin{figure}[!hbt]
\centering
    \subfloat[\label{sfig:sl_fs_1}]{
        \includegraphics[width=.45\textwidth]{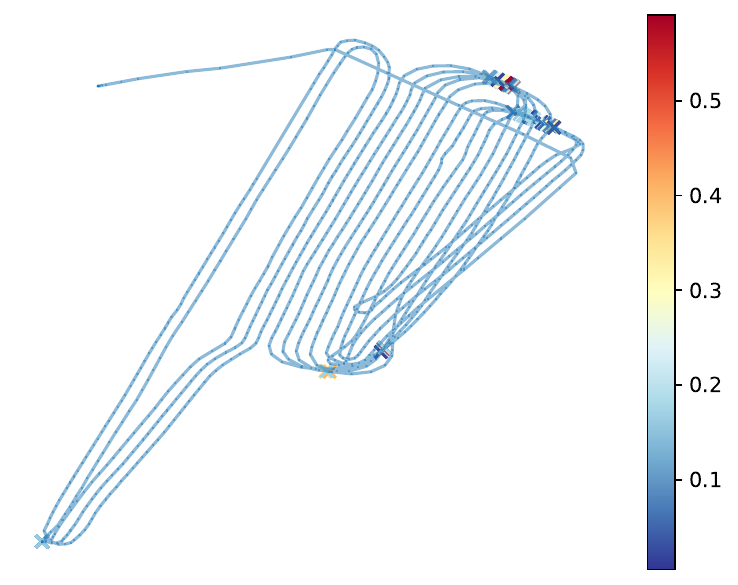}
    }
    \hfill
    \subfloat[\label{sfig:sl_fs_2}]{
        \includegraphics[width=.45\textwidth]{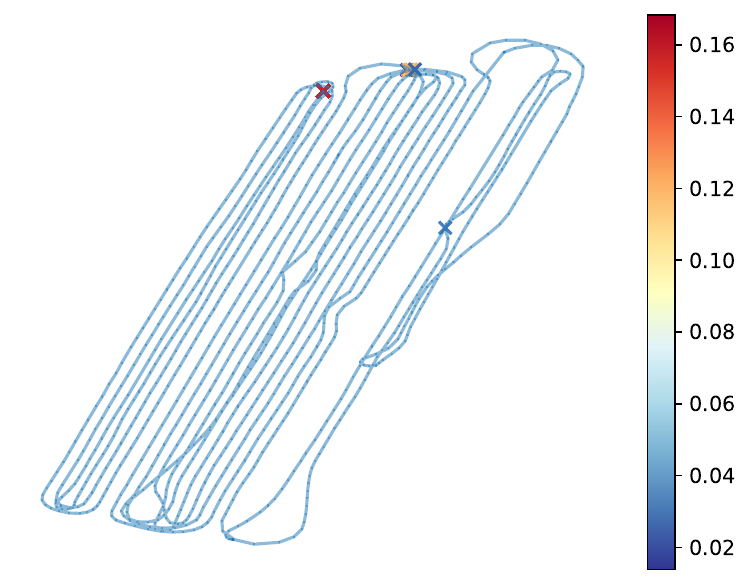}
    }
    \hfill
    \subfloat[\label{sfig:sl_fs_3}]{
        \includegraphics[width=.45\textwidth]{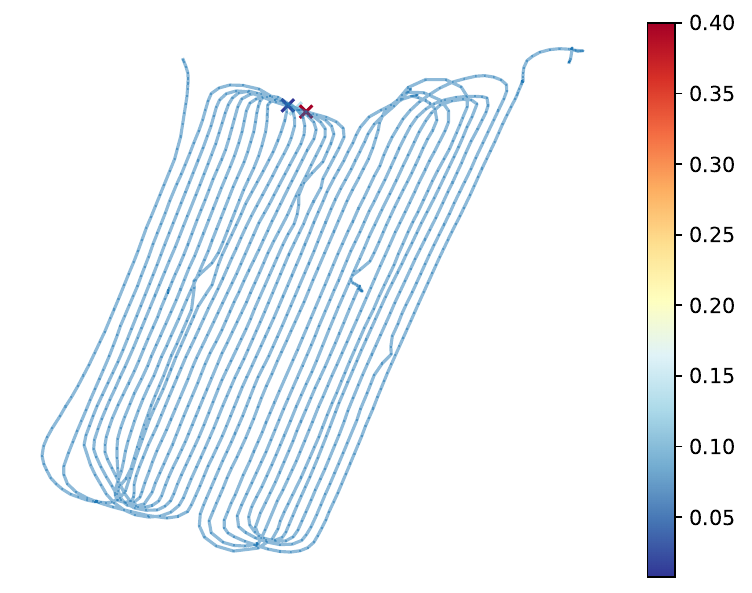}
    }
    \hfill
    \caption{Resulting loop detections overlaid on the trajectories for the FieldSAFE dataset sessions. Crosses stand for detected loops, and their colors represent the absolute distance error between the estimated and the ground truth relative pose.}
    \label{fig:sloopdetector_maps}
\end{figure}

The loop detections are shown overlaid in the respective trajectory maps in Fig.~\ref{fig:sloopdetector_maps}.
It can be seen that most loop detections are found grouped on several groups and on the periphery of the trajectories, which we attribute to the presence of near features that can be easily triangulated as we will discuss later when we show some example images.

\begin{figure}[!hbt]
    \centering
    \subfloat[\label{sfig:sl_error_boxplots}]{
        \includegraphics[width=.48\textwidth]{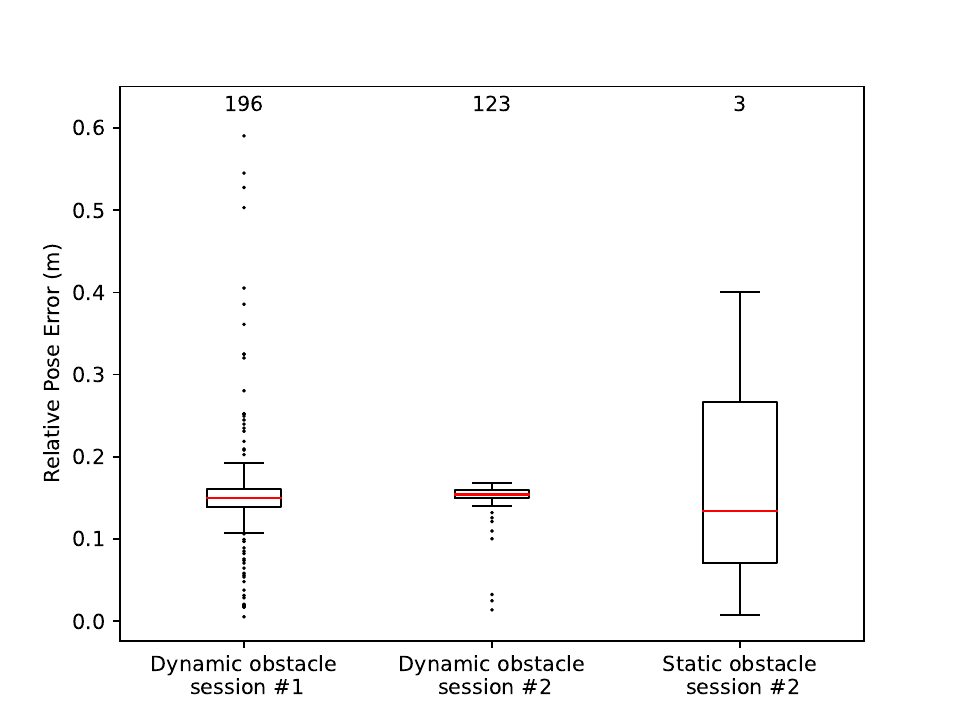}
    }
    \hfill
    \subfloat[\label{sfig:sl_distance_boxplots}]{
        \includegraphics[width=.48\textwidth]{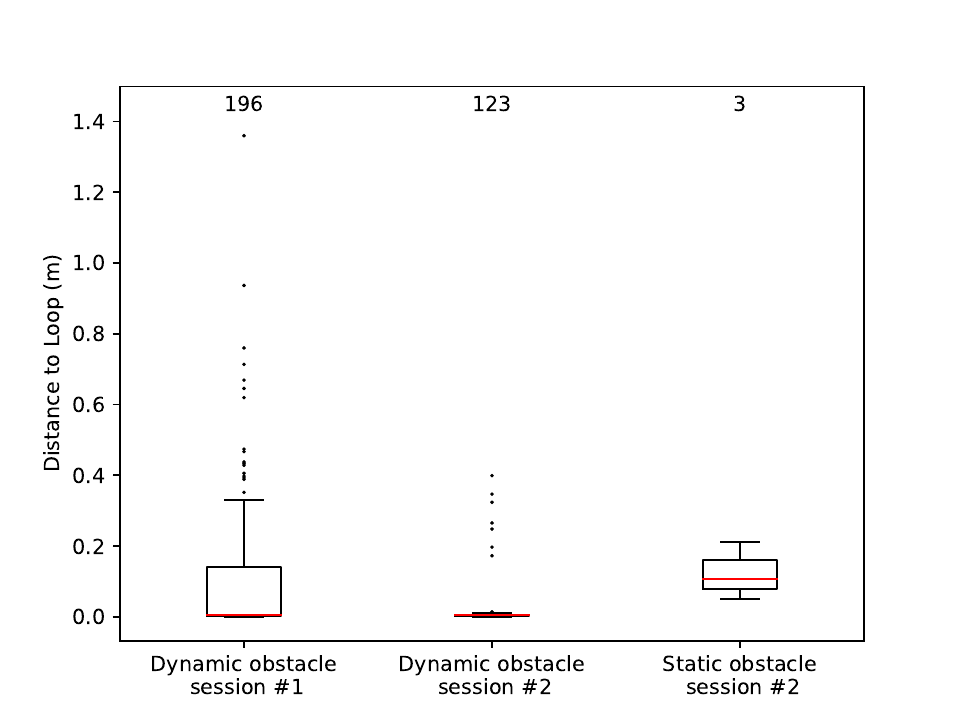}
    }
    \caption{Image~\ref{sfig:sl_error_boxplots} shows the distribution (as boxplots) of the absolute error between distances predicted for the loop detections by our method and actual distance given by each dataset's ground-truth. Image~\ref{sfig:sl_distance_boxplots} shows the distribution of the distance to the loop from each dataset's ground-truth. The number on top of each boxplot shows how many samples are taken into account on computing each boxplot (equal to the number of loops detected in each sequence).}
    \label{fig:sloopdetector_boxplot_metrics}
\end{figure}

Fig.~\ref{fig:sloopdetector_boxplot_metrics} shows the distribution of the absolute errors between the relative pose predicted by our method and the relative pose given by the ground-truth.
We see that our method is able to achieve remarkable results on loop detection well below the $\SI{1}{\meter}$ (arbitrary) error threshold.
It also shows a summary of the ground-truth distances from the pose to its detected loop, in which it can be seen that our method is able to achieve loop detection only for poses with distances of at most $\SI{1.5}{\meter}$, but mostly when the vehicle is traversing almost on the exact same place.

\begin{figure}[!hbt]
\centering
    \subfloat[\label{sfig:sl_fs_1_samplefourmatch}]{
        \includegraphics[width=.48\textwidth]{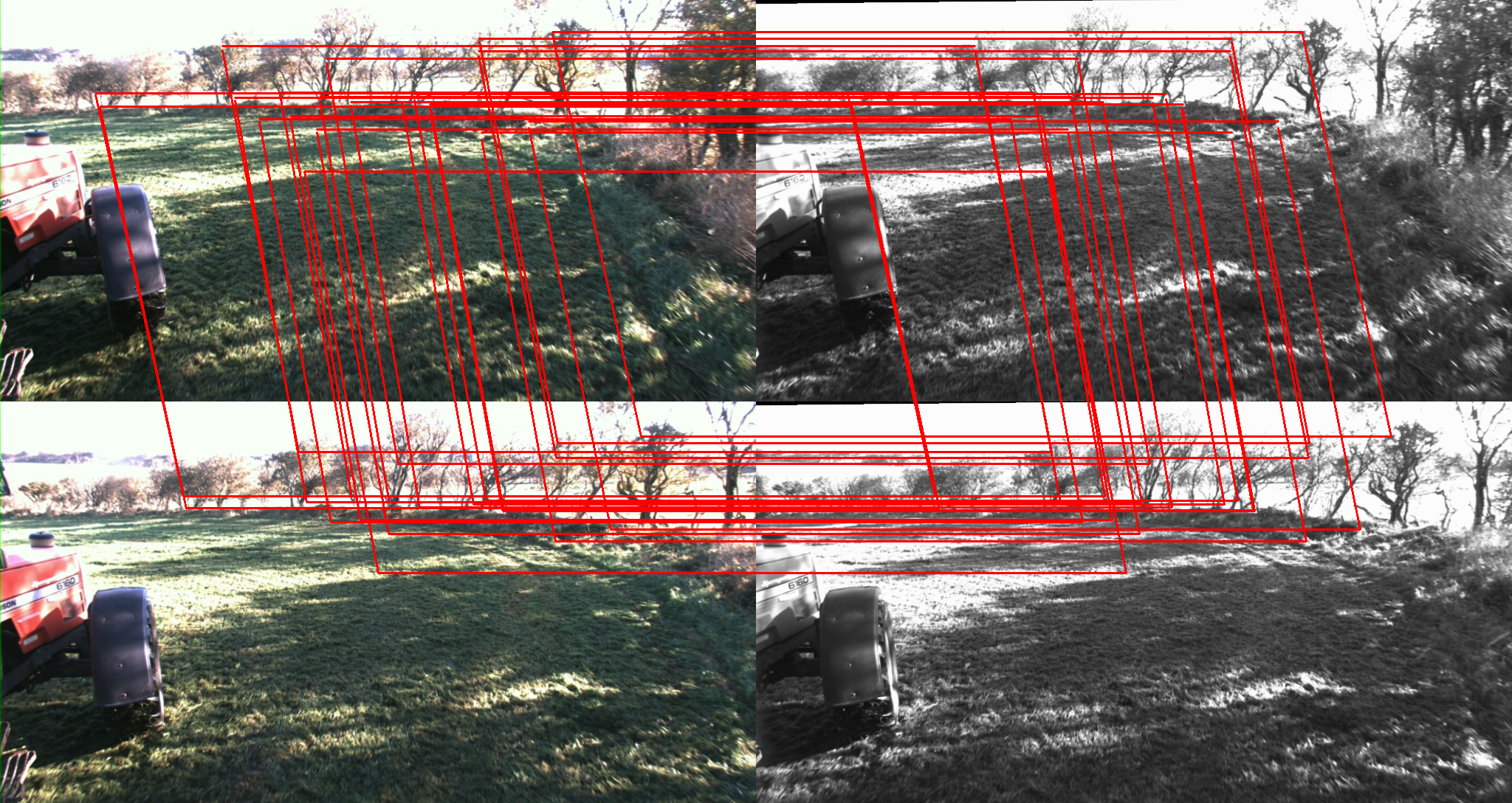}
    }
    \hfill
    \subfloat[\label{sfig:sl_fs_2_samplefourmatch}]{
        \includegraphics[width=.48\textwidth]{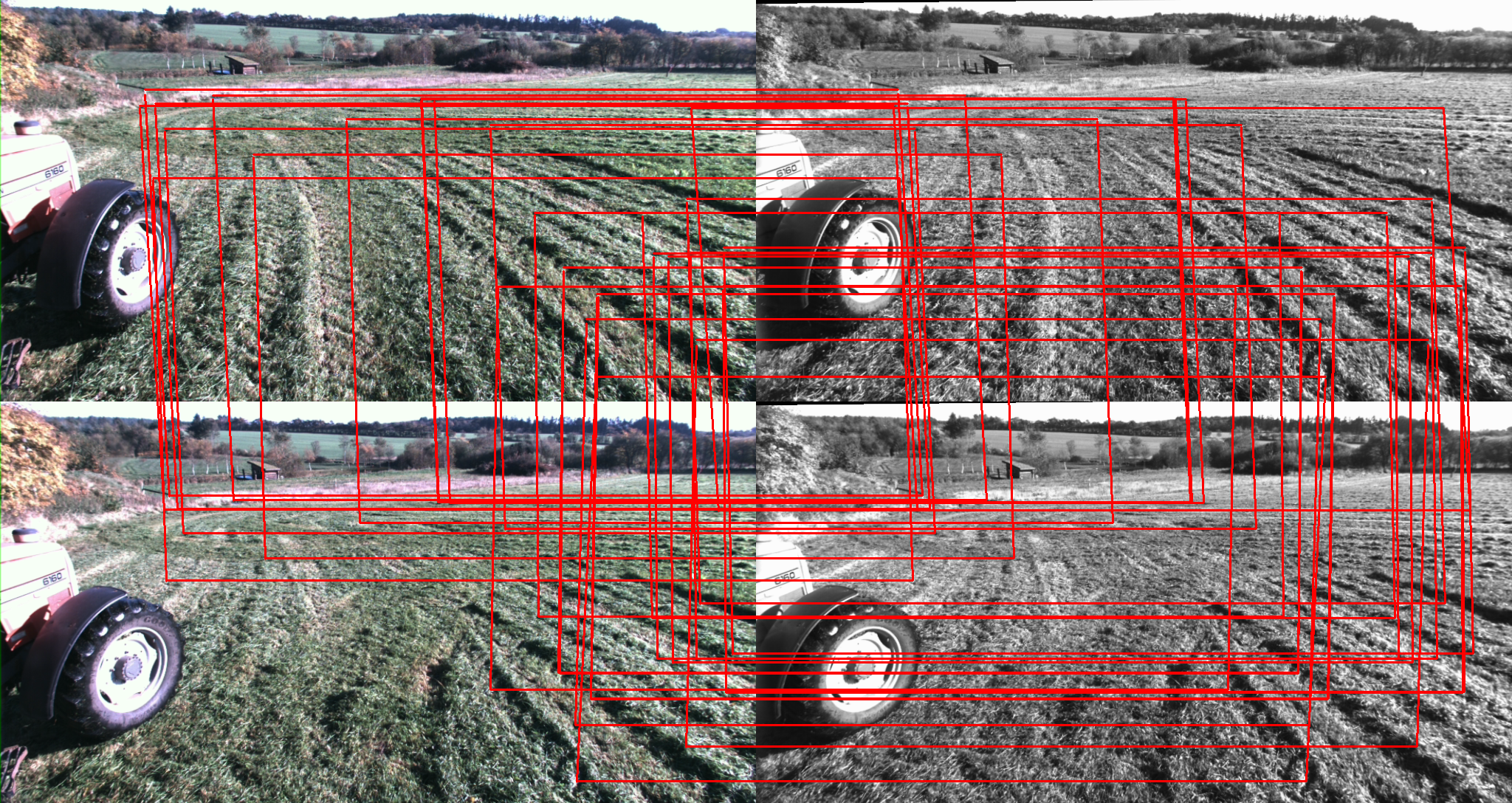}
    }
    \\
    \subfloat[\label{sfig:sl_fs_3_samplefourmatch}]{
        \includegraphics[width=.48\textwidth]{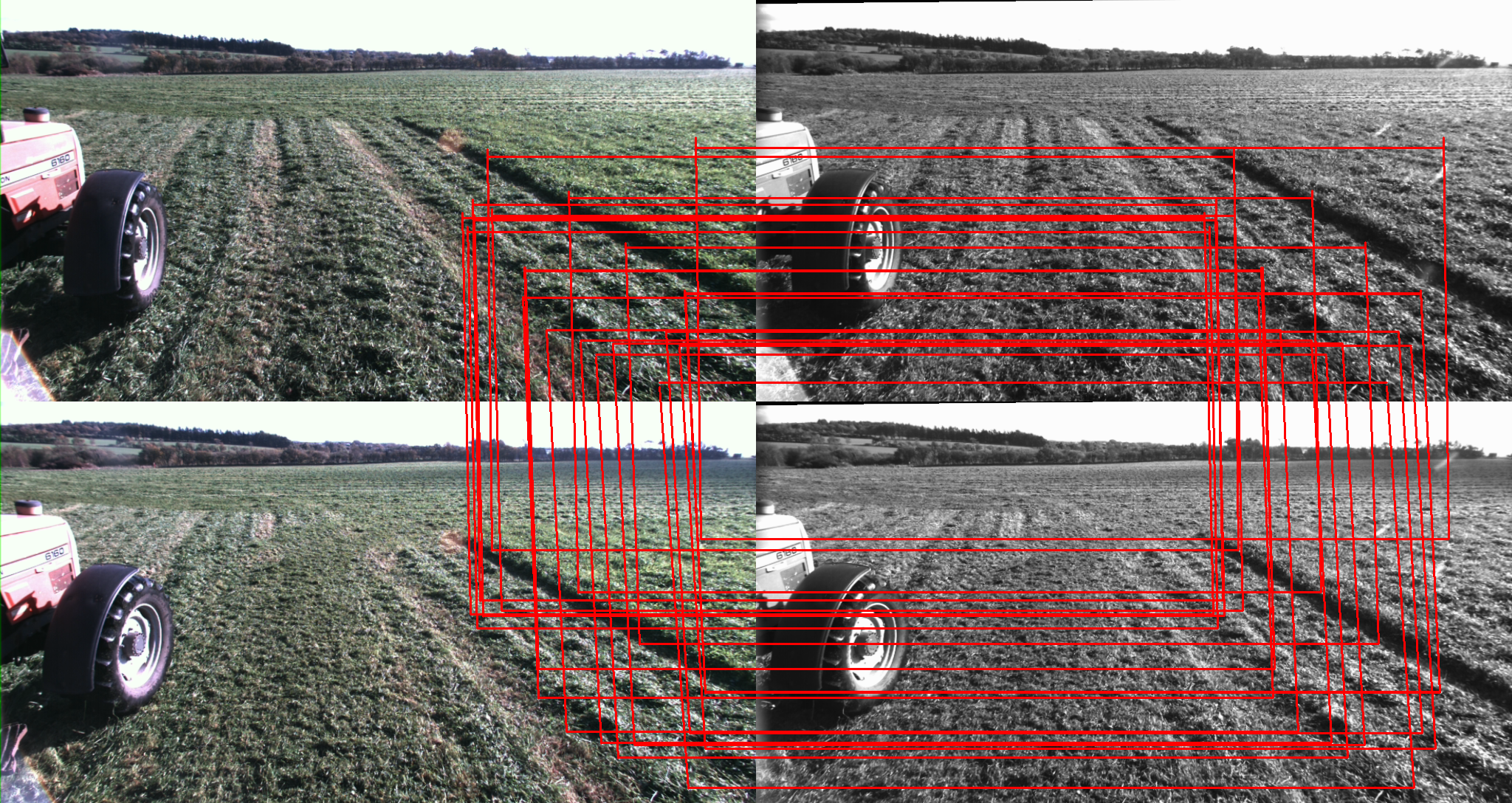}
    }
    \caption{One example for each session of correct loop detections from our stereo matching system, in a two by two tile to accomodate both stereo images, with lines connecting the keypoints used for the final relative pose estimation. Note that the right stereo images are grayscale, as that is what the camera used in the dataset provides.}
    \label{fig:field_detection_images}
\end{figure}

Analysis of these results show that loop detections are achieved on the edges of the field, which can be attributed to the closeness of the features, leading to the successful triangulation of keypoints and relative position regression.
Examples of this can be seen on Fig.~\ref{fig:field_detection_images}, one for each session.
The first sample detection can be seen to have used the shadow of a line of trees, which acts as a landmark on short sequences such as these.
The second sample detection shows nearby distinctive trees and buildings that can be used to get a good estimate of the relative position.
Finally the third sample shows that nearby features on the ground can also make for distinctive elements when calculating the relative position.

\subsection{Analysis and Discussion}

\subsubsection{Visual Place Recognition Analysis}
\begin{figure}[!hbt]
\centering
    \subfloat[\label{sfig:db_fs_1}]{
        \includegraphics[width=.32\textwidth]{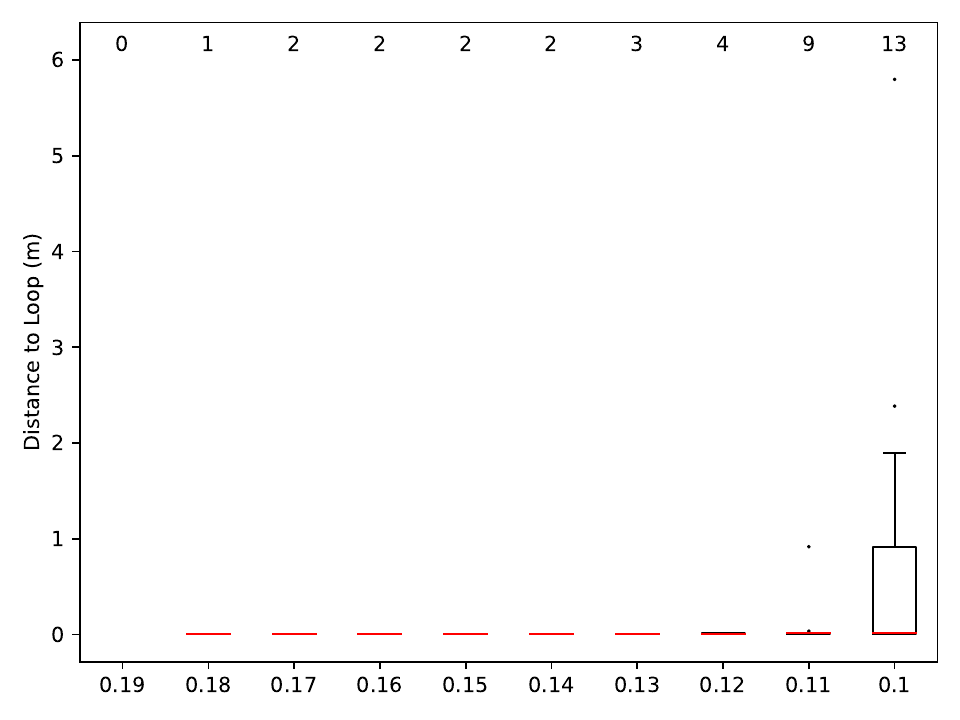}
    }
    \hfill
    \subfloat[\label{sfig:db_fs_2}]{
        \includegraphics[width=.32\textwidth]{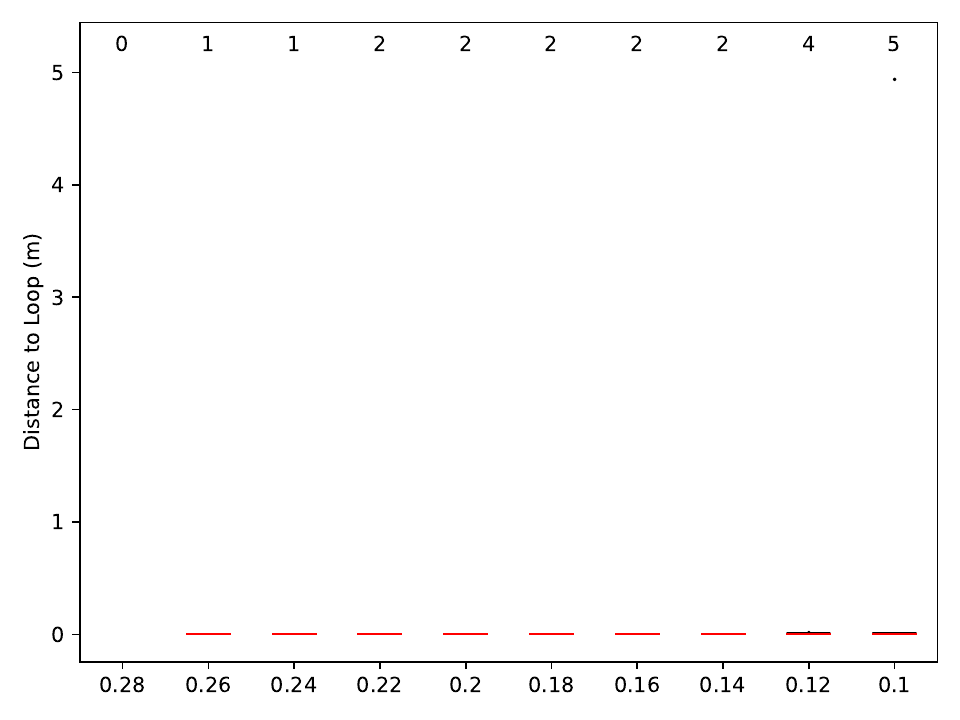}
    }
    \hfill
    \subfloat[\label{sfig:db_fs_3}]{
        \includegraphics[width=.32\textwidth]{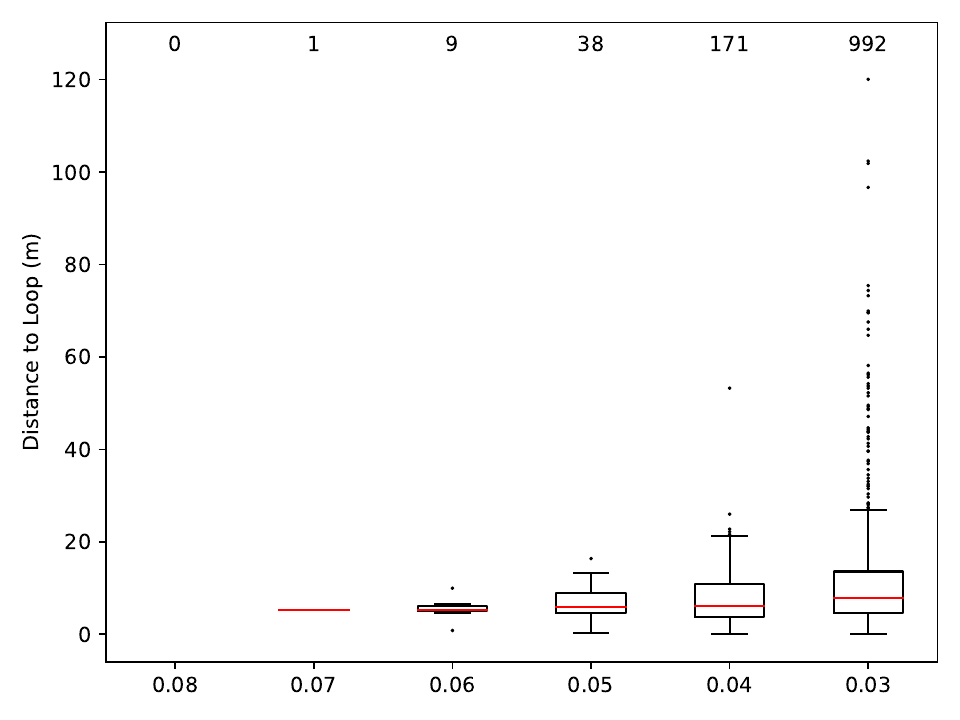}
    }
    \\
    \subfloat[\label{sfig:nv_fs_1}]{
        \includegraphics[width=.32\textwidth]{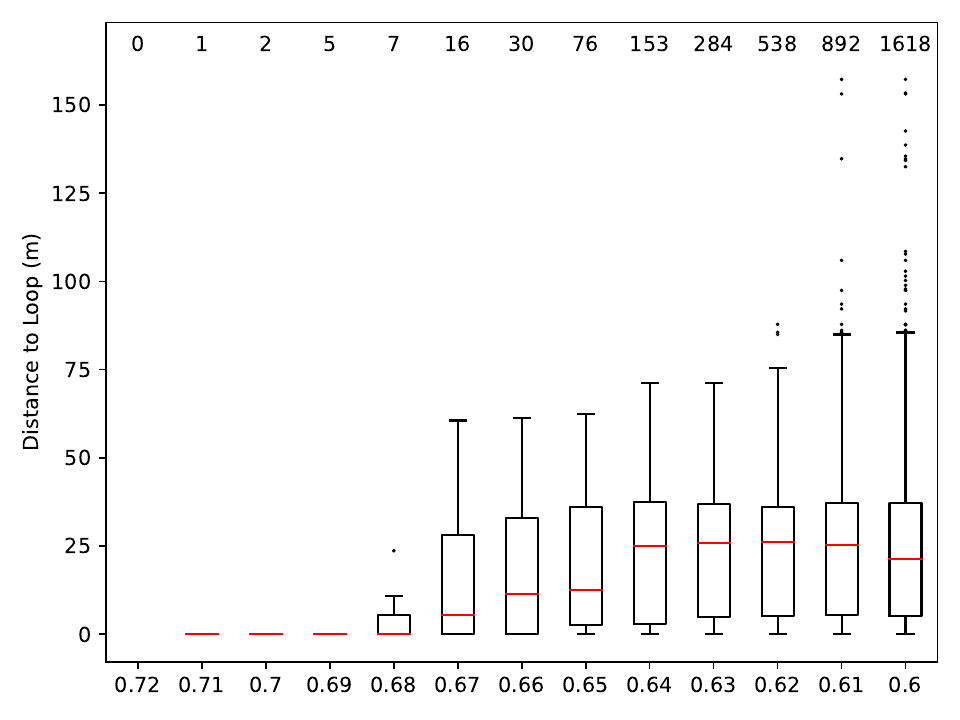}
    }
    \hfill
    \subfloat[\label{sfig:nv_fs_2}]{
        \includegraphics[width=.32\textwidth]{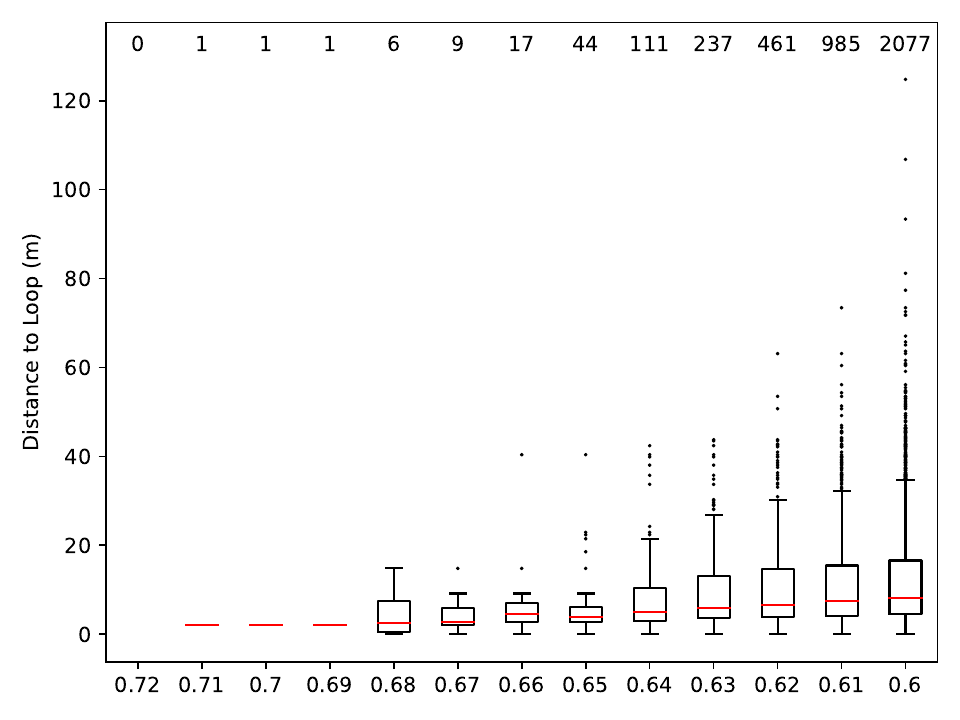}
    }
    \hfill
    \subfloat[\label{sfig:nv_fs_3}]{
        \includegraphics[width=.32\textwidth]{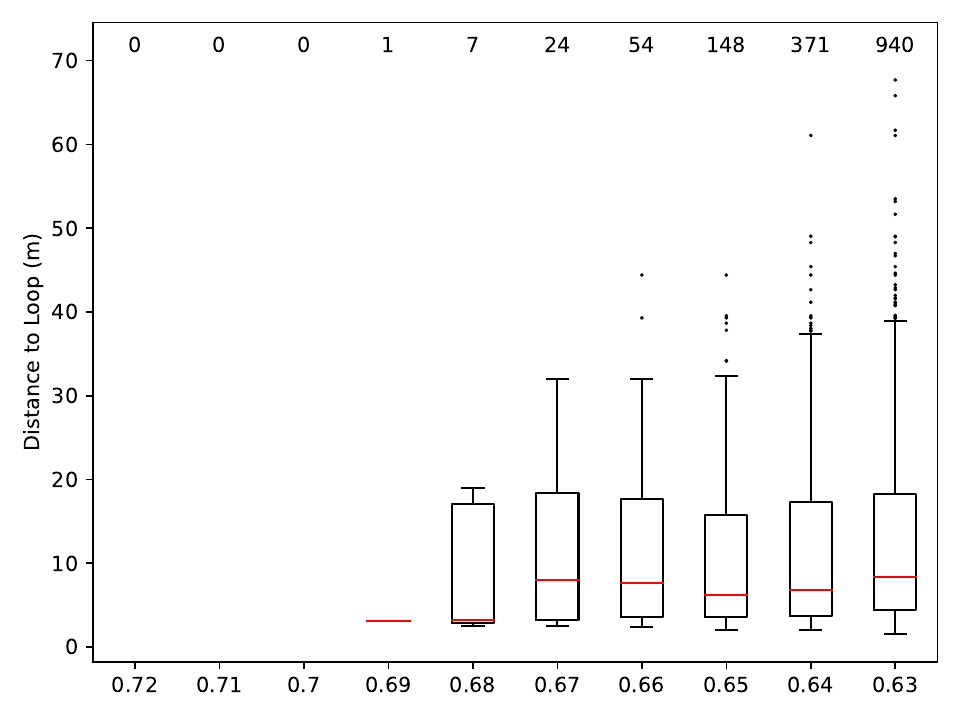}
    }
    \caption{Results obtained by running the VPR systems on the FieldSAFE processed dataset, summarized in boxplots over the ground-truth distance of the matching images for different thresholds. Images \ref{sfig:db_fs_1}, \ref{sfig:db_fs_2} and \ref{sfig:db_fs_3} show results for DBoW2, while images \ref{sfig:nv_fs_1}, \ref{sfig:nv_fs_2} and \ref{sfig:nv_fs_3} show results for NetVLAD. The number on top of each boxplot shows how many samples are taken into account on computing each boxplot. In DBoW2 results the x axis has different scales in the different graphs as it was very difficult to show results for such varying similarity thresholds and resulting matches.}
    \label{fig:vpr_original_results}
\end{figure}

We analyzed how Visual Place Recongition performs on loop detection as standalone and also as a starting point to incorporate it in our method. 
For that we ran some standard experiments on four VPR methods mentioned beforehand: DBoW2~\cite{dbow2012galvezlopez}, NetVLAD~\cite{arandjelovic2015netvlad}.
Since these methods compare all images of the dataset with all other images they can be quite time consuming to run, thus we imitate the keyframe selection common to SLAM systems, whereby we filter the image from the dataset when the system has moved a certain distance from where the last image was selected, in particular we do this for every \SI{0.5}{\meter} traveled or \SI{10}{\degree} of rotation, whichever happens first.
In our experiments we use the position and orientation from the ground-truth GPS systems to select these keyframes, although we understand that a final Loop Detector system would determine these by other means such as visual dissimilarity, odometry, input commands, and/or position estimate.
The experiments consist on running these selected VPR methods on each image pair $(I_i, I_j) \in I$ and obtain a similarity value $s(I_i, I_j) = s_{ij}$, and then evaluate the results sampling through several similarity values as possible thresholds to filter good results from bad results.
We plot the results as boxplots showing the distribution of the distances between the image pairs whose similarity values are over the thresholds.
The resulting distributions, as can be seen in Fig.\ref{fig:vpr_original_results}, show a clear image that there exists no threshold that can both satisfy a high amount of matches and a close distance between them for any method, and that is independent of the specific session of the dataset.

\begin{figure}[!hbt]
    \centering
    \subfloat[\label{sfig:urban_map}]{
        \includegraphics[width=.48\textwidth]{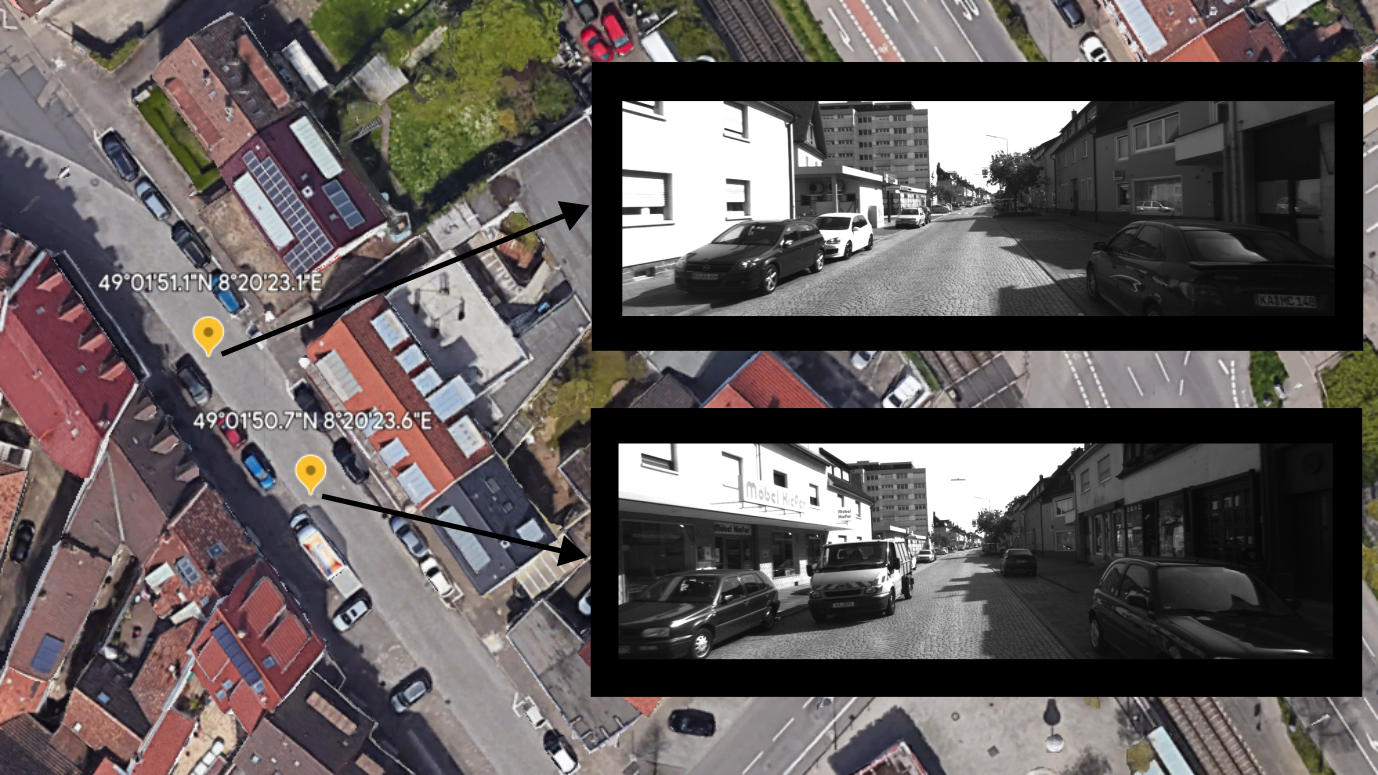}
    }
    \hfill
    \subfloat[\label{sfig:agricultural_map}]{
        \includegraphics[width=.48\textwidth]{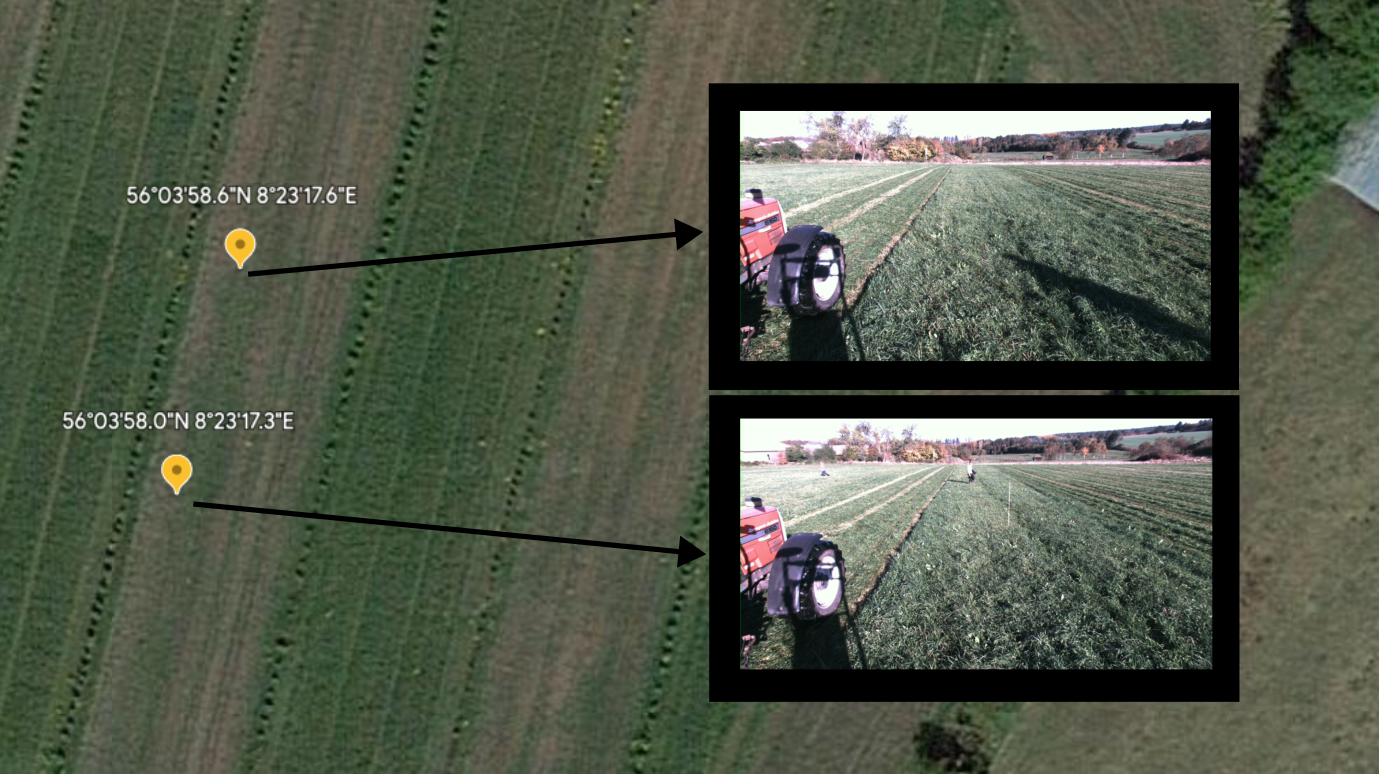}
    }
    \caption{Comparison of a pair of images of the KITTI dataset {Germany} and another from the FieldSAFE dataset {Denmark}. It exemplifies how images taken at the same distance (in this case $\sim{\SI{17}{\meter}}$) in urban environments \ref{sfig:urban_map} present high visual distinctiveness, while those taken in agricultural environments \ref{sfig:agricultural_map} present high visual similarity.}
    \label{fig:urban_vs_agricultural}
\end{figure}

On further review of the results we found a compelling argument as to why a loop detection system based solely on visual information would not produce good results in agricultural or open field environments, when it does on urban or structured ones.
When comparing two images from the FieldSAFE\cite{kragh2017fieldsafe} dataset and contrasting them with those of an urban dataset such as KITTI\cite{geiger2013kittidataset}, as can be seen in Fig.\ref{fig:urban_vs_agricultural}, we appreciate that the visual footprint changes very slightly if at all, which is related to our characterization of agricultural environments as perceptually aliased and could explain the high values on distances for detected loops in the VPR system results.
It is based on these findings that we guided our selection to a BoW-based system that provides local features, which we can use to effectively triangulate a relative pose between the query image and matching loop detection to try and mitigate some of these issues.

\subsubsection{Ablation Study} \label{sec:ablations}
We performed some additional experiments to understand how different parameters of the system affect the loop detection.

\begin{figure}[!hbt]
    \centering
    \includegraphics[width=.5\textwidth]{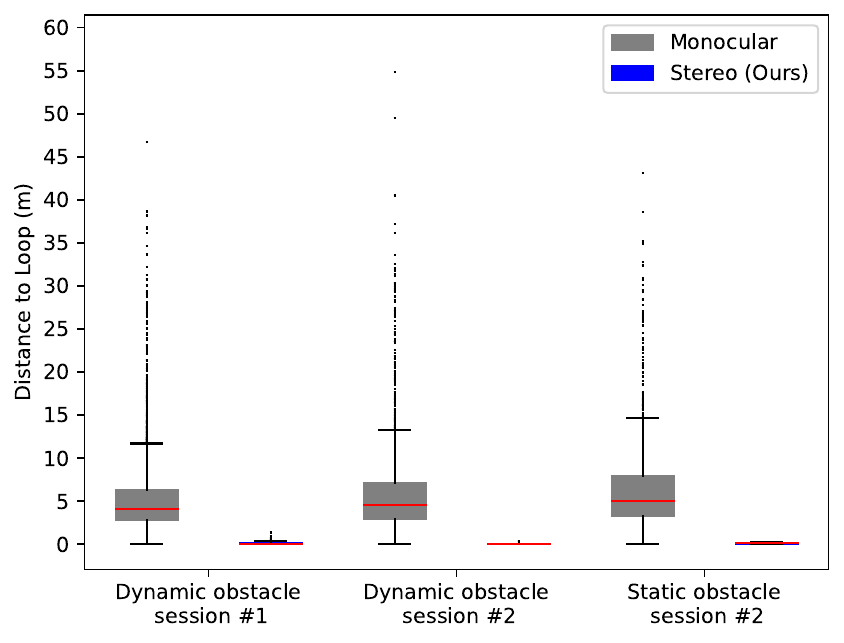}
    \caption{Distribution of the distance to the loop for our proposed stereo system with relative pose estimation and the original monocular system.}
    \label{fig:dloop_vs_sloop_ablation}
\end{figure}

Firstly we compare the results of our stereo loop detection with relative pose estimation to the original system, that is the one presented in \cite{dbow2012galvezlopez}, that works on monocular images and only performs a fundamental matrix check between the images.
Since the original system does not result in a relative position between the query image and the detected loop, we can only compare the absolute distance between the detected loop and the query pose.
The results, as shown in Fig.~\ref{fig:dloop_vs_sloop_ablation} show that the original method results in loops with a much greater distance, and as it doesn't provide an estimate on the relative position a loop closure system based on it would inevitably break the localization and mapping of a SLAM system.

\begin{figure}[!hbt]
    \centering
    \subfloat[\label{sfig:sl_fs_ablation_1}]{
        \includegraphics[width=.48\textwidth]{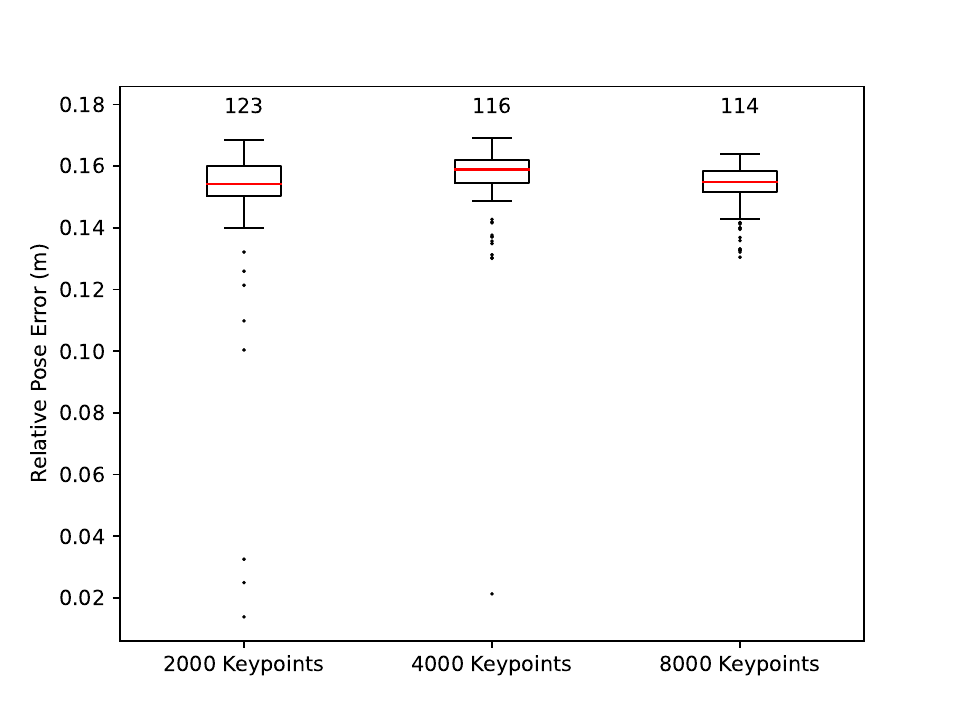}
    }
    \hfill
    \subfloat[\label{sfig:sl_fs_ablation_2}]{
        \includegraphics[width=.48\textwidth]{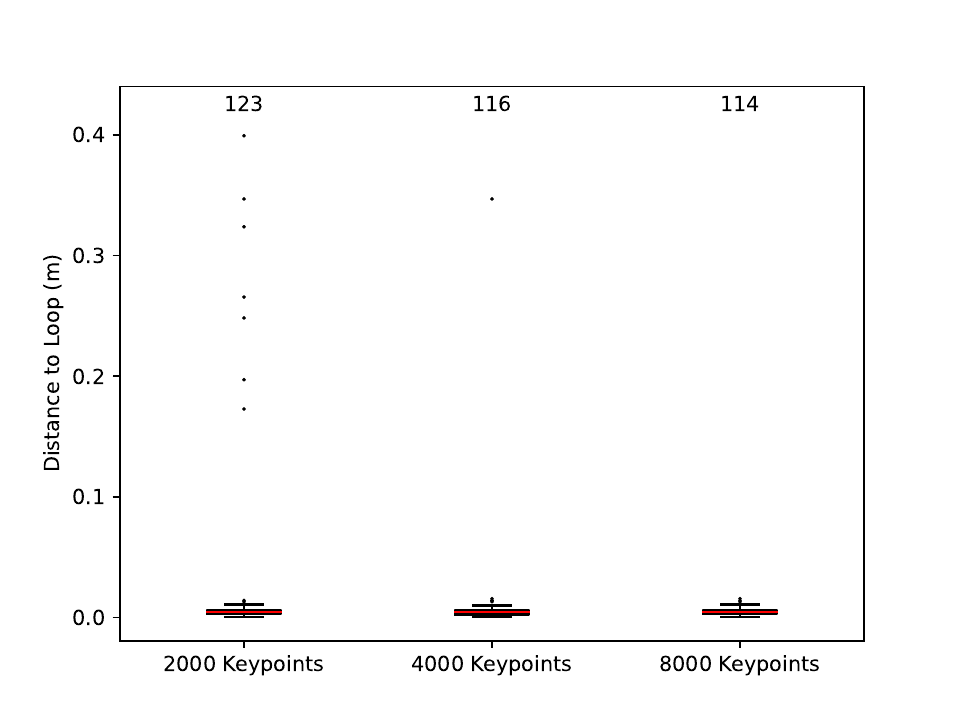}
    }
    \caption{Distributions in the same way as Fig.~\ref{fig:sloopdetector_boxplot_metrics} showing the different metrics for different numbers of ORB keypoints extracted and an amount of minimum points to compute relative pose estimation equal to $1\%$ of this number.}
    \label{fig:sloop_nfeatures_ablation}
\end{figure}

Then we modified the number of features detected by the initial feature detector, always working with ORB, and modified the threshold for number of features needed for the relative pose estimate accordingly, as $1\%$ of the total number of features.
For example for $4000$ keypoints extracted we set the minimum number of points to compute the relative pose estimation between the query image and the matching image at $40$ points.
The results, as shown in Fig.~\ref{fig:sloop_nfeatures_ablation}, show that increasing the number of keypoints extracted per image decreases the number of matches found and makes it so that the loops detected have to be much closer to the actual original pose, with no improvement in maximum relative pose estimation error.
Not shown in the graphics is also an increase in the average compute time for extracting the keypoints of $\SI{45}{\ms}$ for the $2000$ keypoints run, $\SI{58}{\ms}$ for the $4000$ one and $\SI{83}{\ms}$ for the final $8000$ one.
In the same manner the average compute time for the loop detection stages went from $\SI{5}{\ms}$ on the $2000$ keypoints run, to $\SI{20}{\ms}$ on the $4000$ one and then to $\SI{77}{\ms}$ on the $8000$ keypoint run.
It is clear from these numbers that increasing the keypoints extracted per image to such magnitudes would not be feasible for running this loop detection stage in a real time SLAM system.

\section{Conclusions and Future Work} \label{sec:conclusions}
In this paper we present a loop detection method based on stereo vision capable to work on agricultural environments. 
We show that the relative pose regression step is very important in filtering out wrong detections, and can provide valuable information for a following loop closure stage while performing robustly on these environments.
We also explore agricultural environments as what we believe to be a novel class of environments for loop detection and we pave the way to continue researching them in the spirit of achieving robust localization in such places. 
The proposed method was validated in a public agriculture dataset. The results show that the method is able to detect loops and compute the relative pose to it with a median error of $\SI{15}{\cm}$.

Exploring the performance of localization systems in agricultural environments present quite a few new challenges, such as the high degree of visual similarity.
Our results show that using near features is key to computing the loop's relative pose in the geometrical consistency check.
However, since near distinctive features are relatively scarce and changing in these environments, a next step is to take advantage of the features at the horizon to continue loop detection even in the cases where the former cannot be relied upon.

Another important step is to strengthen the different parts of our system to provide us with insight on how bad the estimation is by incorporating uncertainty propagation, from the feature detection stage all the way to the final relative pose estimation. 

Finally, trained features have found recent relevance on keypoint extraction, and they could be useful for finding discriminative features in the fields, such as \cite{vakhitov2021uncertaintyawareestimation}.

\section*{acknowledgements}
This work was partially supported by Consejo Nacional de Investigaciones Científicas y Técnicas (Argentina) under grants PIBAA No.0042, AGENCIA I+D+i (PICT 2021‐570), and by Universidad Nacional de Rosario (PCCT‐UNR 80020220600072UR). This work was also supported by the Spanish Government (PID2021127685NB‐I00 and TED2021‐131150B‐I00) and the Aragón Government (DGA T45\_23R).

\printendnotes

\bibliography{bibliography}

\begin{biography}[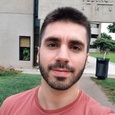]{Nicolás Soncini}
Nicolas has a Computer Science Master and is a PhD student at the CIFASIS (UNR-CONICET) institution under the supervision of Taihú Pire and Javier Civera. He has experience working in computer vision in microcomputers and in consumer robotics, and his work aims toward improving vision for autonomous robotics in agricultural settings, particularly in loop detection and loop closing for SLAM systems.
\end{biography}

\begin{biography}[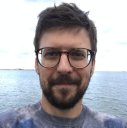]{Javier Civera Sancho}
Javier Civera is a Professor at the University of Zaragoza in Spain, where he is in charge of machine learning, computer vision and SLAM courses and leads a research group focused on 3D computer vision and visual SLAM. He is the co-author of over 80 scientific publications on the topic, cited over 8,200 times according to Google Scholar. He has led several  research and transfer projects and serves currently as Editor for IEEE Transactions on Robotics and IEEE Robotics and Automation Letters.
\end{biography}

\begin{biography}[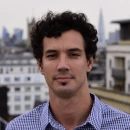]{Taihú Aguara Nahuel Pire}
Taihú Pire was born in Rosario, Argentina, in 1986. He received the licentiate degree in Computer Science (2010) at the National University of Rosario and the PhD in Computer Science (2017) at the University of Buenos Aires. He is the head of the robotics lab at CIFASIS institute at the National Science and Technology Council of Argentina and Adjunct Professor at National University of Rosario. Currently, his research interests are in developing new SLAM algorithms and autonomous navigation systems.
\end{biography}

\end{document}